\documentclass[runningheads]{llncs}
\usepackage{graphicx}
\usepackage[hang,flushmargin]{footmisc}
\usepackage[misc,geometry]{ifsym}
\usepackage{cite} 
\usepackage[T1]{fontenc}
\usepackage{amsmath}
\usepackage{mathrsfs}
\usepackage{graphicx}  
\usepackage{epstopdf}
\usepackage{bbm}
\usepackage{float}
\usepackage{multirow}
\usepackage{booktabs}
\usepackage{mathrsfs}
\usepackage{amssymb}
\usepackage[colorlinks,linkcolor=red]{hyperref}
\usepackage{siunitx}
\makeatletter

\newcommand{\Rmnum}[1]{\expandafter\@slowromancap\romannumeral #1@}
\makeatother
\usepackage{booktabs}
\usepackage{amsfonts,amssymb}
\usepackage{amsmath,bm}
\usepackage{array}
\newcommand{\PreserveBackslash}[1]{\let\temp=\\#1\let\\=\temp}
\newcolumntype{C}[1]{>{\PreserveBackslash\centering}p{#1}}
\newcolumntype{R}[1]{>{\PreserveBackslash\raggedleft}p{#1}}
\newcolumntype{L}[1]{>{\PreserveBackslash\raggedright}p{#1}}
\usepackage{algorithmicx,algorithm}
\makeatletter

\usepackage{xcolor}
\begin{document}
\title{Cheap Lunch for Medical Image Segmentation by Fine-tuning SAM on Few Exemplars}
\titlerunning{Cheap Lunch for Medical Image Segmentation}
\author{
Weijia Feng\inst{1}\and
Lingting Zhu\inst{2}\and
Lequan Yu\inst{2}\textsuperscript{(\Letter)}
}
%index{Feng, Weijia}
%index{Zhu, Lingting}
%index{Yu, Lequan}

\authorrunning{W. Feng et al.}
\institute{
    Zhejiang University, Hangzhou, China \\ \email{22135081@zju.edu.cn} \and
    The University of Hong Kong, Hong Kong SAR, China \\ \email{ltzhu99@connect.hku.hk lqyu@hku.hk}
}
\maketitle

\begin{abstract}
The Segment Anything Model (SAM) has demonstrated remarkable capabilities of scaled-up segmentation models, enabling zero-shot generalization across a variety of domains.
By leveraging large-scale foundational models as pre-trained models, it is a natural progression to fine-tune SAM for specific domains to further enhance performances. 
However, the adoption of foundational models in the medical domain presents a challenge due to the difficulty and expense of labeling sufficient data for adaptation within hospital systems.
In this paper, we introduce an efficient and practical approach for fine-tuning SAM using a limited number of exemplars, making it suitable for such scenarios. Our approach combines two established techniques from the literature: an exemplar-guided synthesis module and the widely recognized Low-Rank Adaptation (LoRA) fine-tuning strategy, serving as data-level and model-level attempts respectively.
Interestingly, our empirical findings suggest that SAM can be effectively aligned within the medical domain even with few labeled data.
We validate our approach through experiments on brain tumor segmentation (BraTS) and multi-organ CT segmentation (Synapse). The comprehensive results underscore the feasibility and effectiveness of such an approach, paving the way for the practical application of SAM in the medical domain. 
\keywords{Medical Image Segmentation  \and Foundation Models \and Segment Anything Model (SAM) \and Few Exemplars}
\end{abstract}
\section{Introduction}

Nowadays, foundation models~\cite{bommasani2021opportunities} have revolutionized the AI community, demonstrating immense potential to solve tasks within an integrated framework and achieve remarkable zero-shot and few-shot performances~\cite{brown2020language,chowdhery2022palm}.
The Segment Anything Model (SAM)~\cite{kirillov2023segment}, a promptable model trained on over 1 billion masks and 11 million images, makes an attempt to build a foundation model for segmentation. SAM has shown impressive zero-shot segmentation ability on new data across different distributions and tasks.
However, SAM's performance has been found to be limited in certain domains, such as medical images segmentation~\cite{chen2023sam,huang2023segment,zhang2023customized}, low-level structural segmentation~\cite{chen2023sam}, and intricate objection segmentation~\cite{ke2023segment}. To address these limitations, researchers have sought to enhance the performance of pre-trained models across domains by fine-tuning SAM or externally designed components~\cite{chen2023sam,ke2023segment,ma2023segment,zhang2023customized}. As a result, fine-tuning SAM with medical images could be more feasible and promising to facilitate segmentation tasks in real clinical applications~\cite{zhang2023customized}.

Despite these advancements, the adoption of medical segmentation in real hospitals remains challenging due to the need for large curated datasets. Fine-tuning SAM on labeled images of specific instruments is also required to align model's understanding of the domain scope within the hospital. This introduces a time-consuming, labor-intensive, and expensive process of data labeling~\cite{wang2021annotation}. Consequently, there is growing interest in developing effective methods to leverage limited annotated data for training deep learning models~\cite{chaitanya2020contrastive,en2022exemplar,wang2021annotation}.

Among the various attempts to utilize small sets of labeled data, we consider exemplar-based learning an intriguing approach. This scenario, which involves using a single expert-annotated image that covers all parts of the whole organ category set~\cite{en2022exemplar}, can significantly reduce the labeling expenses in hospital systems. This raises the question: \textbf{Can we fine-tune foundation models (SAM) on few exemplars to achieve significant improvements in medical image segmentation?} 

In this paper, we integrate two well-established techniques from the literature to serve as the data-level and model-level attempts. On the data-level, we employ the exemplar-guided synthesis module in~\cite{en2022exemplar} to generate a synthetic training dataset through geometric and intensity transformation. On the model-level, our fine-tuning strategy is based on the widely recognized Low-Rank Adaptation (LoRA)~\cite{hu2021lora} and specifically, we adhere to the basic architecture outlined in~\cite{zhang2023customized}. Notably, we configure the ViT-Base image encoder and update a total of 6.32 million parameters. Unlike many works relying on A100 40/80G GPUs, all of our experiments are executable on more accessible GPUs such as the 3090 24G GPUs.

We assess the effectiveness of our approach on two medical image segmentation tasks: brain tumor segmentation (BraTS 2018~\cite{bakas2017advancing,bakas2018identifying,menze2014multimodal}) and multi-organ CT segmentation (Synapse\footnote{\url{https://www.synapse.org/\#!Synapse:syn3193805/wiki/217789}}). Extensive results suggest that fine-tuning SAM on a few exemplars can strike a balance between accuracy and annotation labor, offering a cost-effective solution for medical image segmentation.
In summary, our contributions are twofold: (1) We introduce the attempt of fine-tuning the foundation segmentation model SAM with few exemplars for medical image segmentation. (2) We present comprehensive results on two datasets from different sub-domains, using only 1\% labeled data, demonstrating the feasibility of this cost-effective solution. 
\section{Methods}
\subsection{Synthesized Dataset Based on Exemplars}
Given limited labeled exemplars, to generate substituted training dataset comprising of more data samples, we adopt the exemplar-based synthesis module proposed in~\cite{en2022exemplar} to create more synthesized data. 
For each organ or tumor cropped from exemplars, we apply geometric and intensity transformations including blur, intensity variation, scale, flip and rotation to it before pasting it onto similarly transformed background images. Background images are chosen from slices in training volumes without organs or tumors. Due to the different numbers of the segmentation labels in the two datasets, we use different processes for BraTS 2018 (tumor) and Synapse (multi-organ). The processes can be depicted by the following equations.

For the BraTS 2018, the generating of synthesized images can be described by
\begin{equation}
I_s =  \mathcal{T}(I_{e} \otimes Y_{e}) \quad if\quad \mathcal{T}(I_{e} \otimes Y_{e}) > 0 \quad else\quad \mathcal{T}(I_b),
\label{exemplar_organ}
\end{equation}
where $\mathcal{T}$ denotes the transformation, $I_s$, $I_b$, $I_e$, $Y_e$ indicate the synthesized image, background image, exemplar image and exemplar label respectively. The $0$ in the equation refers to the value of background pixels. The label of synthesized images can be extracted naturally from above operations.

Since exemplars from the Synapse contain multiple organs, we follow the category-wise manner in~\cite{en2022exemplar}. We first apply the above equation to each organ and then merge these organs into the same background image after transformation, which can be described as follows:
\begin{equation}
O_s^{k} = \mathcal{T}(I_{e} \otimes Y_{e}^k),
\end{equation}
\begin{equation}
I_s = Merge(O_s^1,...,O_s^K) \quad if\quad Merge(O_s^1,...,O_s^K) > 0 \quad else \quad \mathcal{T}(I_b),
\label{exemplar_organ}
\end{equation}
where $Y_e^{k}$ and $O_s^k$ refer to the label and the transformed image of the $k$-th organ respectively. The $k$ is in $ \{1, ..., K\}$ and $K$ is the total classes. When pasting the synthesized organs onto the background image, we kept the position of the synthesized organs on the background image roughly consistent with their position in the exemplar.

\subsection{Fine-tuning SAM}
Low-Rank Adaptation (LoRA~\cite{hu2021lora}) is originally proposed as a fine-tune strategy that applies low rank decomposition matrices for Transformer~\cite{vaswani2017attention} based large language model. The basic idea can be summarized as updating the pre-trained weight matrix with a low-rank decomposed bypass matrix. The bypass matrix can be treated as the multiplication of two matrices
$A \in \mathbb{R}^{r \times C_{in}}$ and $B \in \mathbb{R}^{C_{out} \times r}$ with the low-rank constraint, \textit{i.e.,} $r \ll \min(C_{in}, C_{out})$. As a result, given a projection layer $W \in \mathbb{R}^{C_{out} \times C_{in}}$, the updated projection is described as $\hat{W} = W + \Delta W = W + BA$. 

Following the key design in LoRA, the basic operations adopting LoRA in computer vision tasks may involve firstly freezing all parameters in the pre-trained model and then applying trainable bypass matrices as projection layers for transformer blocks. In practice, we follow the basic backbone in SAMed~\cite{zhang2023customized} of using LoRA to fine-tune SAM on medical images. Different from SAMed, we investigate point-based prompting in SAM and fine-tune the mask decoder with the point-based prompt embedding. SAMed utilizes LoRA in the query and value projection layers and it is optional to fine-tune all the parameters or apply LoRA to the lightweight mask decoder of SAM. In our implementation, we observe that applying LoRA to the mask decoder achieves better performance in our setting. In order to incorporate point prompts for training all classes simultaneously, we randomly select one class and randomly sample a point from pixels belong to this class during training. In the end, the mask decoder outputs two classes including a background class and the target class.
We apply LoRA to both the image encoder and the mask decoder with rank $r=4$, and there are only 6.32M trainable parameters, which are 1.77\% of SAM that using ViT-B as image encoder. Our fine-tune strategy only needs to train a small number of parameters and serve as a cost-effective solution.

\section{Experiments}

\subsection{Datasets}
We conduct experiments on two medical datasets including the BraTS 2018~\cite{bakas2017advancing,bakas2018identifying,menze2014multimodal} and the Synapse Dataset.
BraTS 2018 contains 285 MRI scans in the original training split, and we randomly split those scans into 80\% for training and the remaining 20\% for testing. The final train set contains 228 scans with a total of 38340 slices, of which 14662 have tumors with no fewer than 10 pixels.
For this dataset, we use the FLAIR modality to segment the whole tumor like these papers~\cite{ghorbel2022transformer,ma2023segment}. For each image in BraTS, we normalize the pixel values to the range [0, 1] using min-max normalization. 
The Synapse dataset is from MICCAI 2015 Multi-Atlas Abdomen Labeling Challenge, consisting of 30 abdominal CT scans.
Following division settings in TransUNet~\cite{chen2021transunet}, 30 cases in Synapse are divided into 18 training cases and 12 testing cases with eight organs. 18 training cases contain 2212 axial slices in total. All Images in Synapse are clipped to [-125, 275] before min-max normalized into [0, 1].
Dice Similarity Score (DSC) and 95\% Hausdorff Distance (HD95) are used as the evaluation metrics.

\subsection{Exemplar Selection}
Due to the different characteristics of the two datasets, we select exemplars with different strategies. For the BraTS 2018 dataset, since our segmentation target is the whole tumor solely, we randomly select exemplars from all training images that having tumors with a portions of 0.5\%, 1\% and 3\% respectively.
For the Synapse dataset, each slice has a different number of organs. We select 9, 18 and 36 exemplars that have most organs in training volumes since the imbalanced organ distribution can lead to poor segmentation results for some organs. Specifically, for the 9 exemplars case, we select 9 from all training volumes, and then choose one image with the most organs from each of the selected volume as an exemplar. We select one and two images from each training volume for the 18 and 36 exemplars cases respectively.

\subsection{Implementation Details}
For the substitute training dataset synthesis, we ensure that for all different exemplar numbers, the synthesized datasets are of the same size, \textit{i.e.,} 4500 for the BraTS 2018 and 1800 for the the Synapse Dataset.
Cross entropy loss and dice loss are adopted to optimize trainable parameters. The overall loss function can be written as 
$L=\lambda_1 L_{CE}+\lambda_2 L_{Dice},$
where $\lambda_1$ is set to 1 and $\lambda_2$ is set to 0.8 empirically.
We use the AdamW~\cite{loshchilov2017decoupled} optimizer with an initial learning rate of $\alpha=0.001$. Warm up and exponential decay are adopted to schedule the learning rate. All experiments are run on two NVIDIA GeForce RTX 3090 GPUs.

\begin{table*}[ht]
\centering
\caption{Quantitative comparison on the BraTS 2018.}
\scriptsize
\begin{tabular}{c|c|cc}
\toprule
\textbf{Methods} & \textbf{Exemplar Nums}& \textbf{DSC} $\uparrow$ & \textbf{HD} $\downarrow$\\
\midrule
\midrule
SAM (Zero-Shot) & - & 45.29 & 54.74 \\
\midrule
\multirow{4}[1]{*}{SAMed (w/ Data Synthesis)} & 75 (0.5\%) & 82.80 & 28.03 \\
& 150 (1\%) & 82.50 &  43.99 \\
& 450 (3\%) & 85.53 & 17.56 \\
& Total Nums & 85.52 & 31.13 \\
\midrule
\multirow{3}[1]{*}{Ours}
& 75 (0.5\%) & 82.78 & 14.92 \\
& 150 (1\%) & 83.4 & 10.03 \\
& 450 (3\%) & 83.07 & 16.94 \\
\midrule
Full Set (Pseudo Upper Bound) & Total Nums & 85.28 & 7.91 \\
\bottomrule
\end{tabular}

\label{Brats-18}

\vspace{0.6cm}
\centering
\caption{Quantitative comparison on the Synapse multi-organ CT dataset.}
\footnotesize
\resizebox{\textwidth}{!}{
\begin{tabular}{c|c|cc|cccccccc}
\toprule
\textbf{Methods} & \textbf{Exemplar Nums}& \textbf{DSC} $\uparrow$ & \textbf{HD} $\downarrow$ & \textbf{Aorta} & \textbf{Gallbladder}& \textbf{Kidney(L)} & \textbf{Kidney(R)}& \textbf{Liver}& \textbf{Pancreas}& \textbf{Spleen}& \textbf{Stomach} \\
\midrule
\midrule
SAM & - & 74.54 & 40.90 & 88.74 & 40.55 & 87.11 & 82.60 & 88.63 & 53.77 & 83.79 & 71.14 \\
\midrule
Att-UNet~\cite{oktay2018attention} &Total Nums & 77.77 & 36.02 &89.55 & 68.88 & 77.98 & 71.11 & 93.57 & 58.04 & 87.30 & 75.75 \\
\midrule
\multirow{4}[1]{*}{SAMed} & 9 (one per two volumes) & 43.82 & 96.21 & 40.50 & 34.20 & 44.35 & 46.10 & 81.32 & 23.43 & 43.86 & 36.78\\
 & 18 (one per volume) & 55.26 & 75.02 & 45.28 & 50.70 & 58.75 & 62.53 & 87.58 & 31.07 & 72.82 & 33.32 \\
 & 36 (one per volume) & 66.96 & 44.69 & 63.75 & 58.17 & 72.97 & 68.96 & 90.67 & 40.38 & 80.40 & 60.35 \\
& Total Nums & 81.88 & 20.64 & 87.77 & 69.11 & 80.45 & 79.95 & 94.80 & 72.17 & 88.72 & 82.06 \\
\midrule
\multirow{4}[1]{*}{Ours} & 1 (one exemplar) & 75.91 & 21.75 & 84.46 & 49.56 & 83.74 & 84.92 & 88.05 & 56.43 & 89.28 & 70.80 \\
& 9 (one per two volumes) & 79.08 & 21.62 & 88.75 & 55.76 & 88.35 & 84.11 & 89.76 & 61.26 & 91.27 & 73.41 \\
& 18 (one per volume) & 83.04 & 16.84 & 89.18 & 71.33 & 89.20 & 86.46 & 92.55 & 64.20 & 90.52 & 80.84 \\
& 36 (two per volume) & 84.23 & 11.86 & 88.31 & 69.91 & 90.43 & 88.57 & 94.82 & 65.17 & 91.36 & 85.24 \\
\midrule
Full Set & Total Nums & 85.95 & 8.97 & 91.52 & 64.42 & 92.49 & 91.56 & 96.06 & 66.49 & 94.39 & 90.66 \\
\bottomrule
\end{tabular}
}
\label{synapse}
\end{table*}

\begin{figure*}[ht]
\begin{center}
\includegraphics[width=\linewidth]{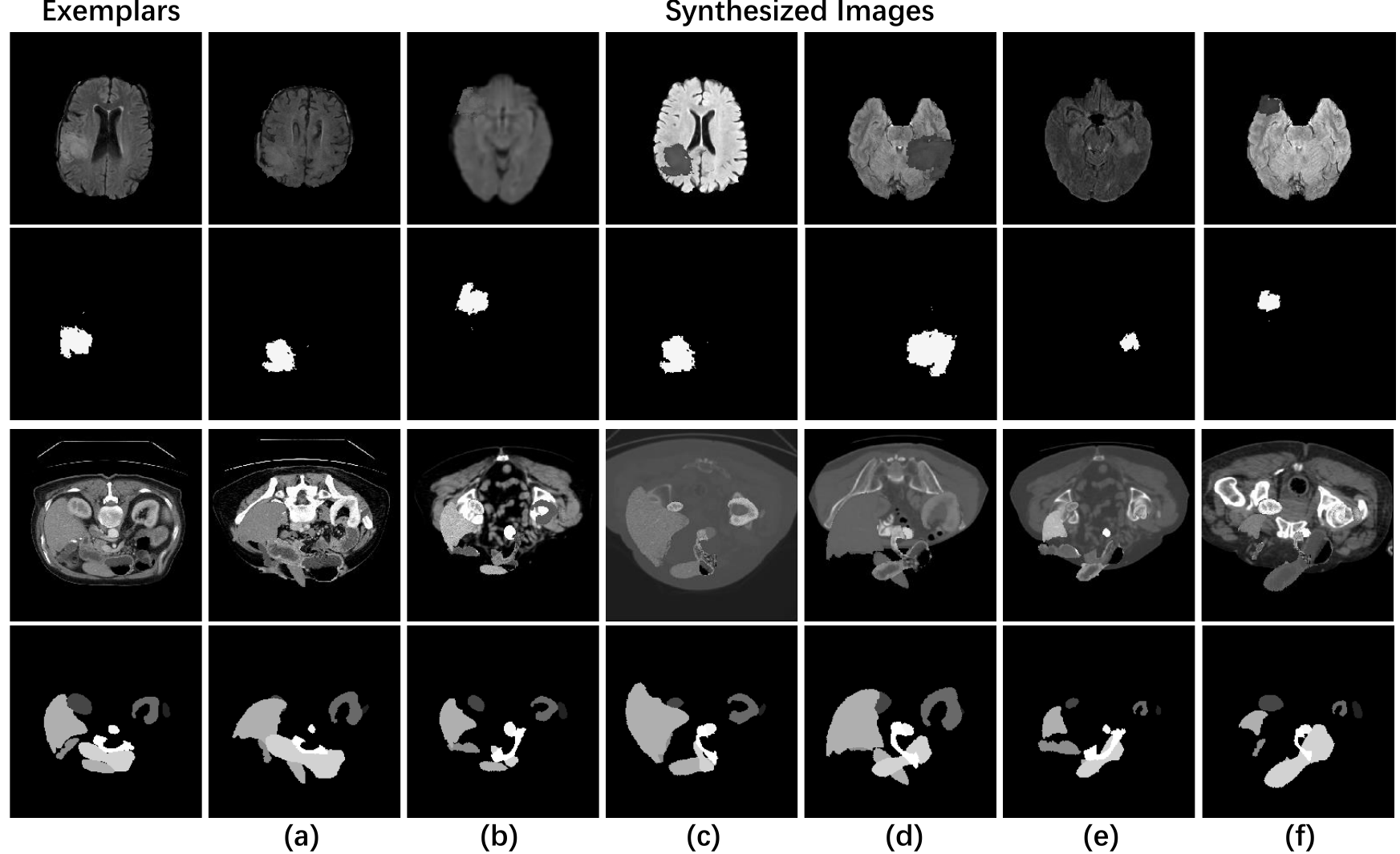}
\end{center}
\vspace{-0.4cm}
\caption{Synthesized data samples. The images are shown in the 1-st and 3-rd rows, while their corresponding labels are presented in the 2-nd and 4-th rows. The first column shows the exemplars, and the subsequent columns present several synthesized data samples. Geometric and intensity transformations including blur, intensity variation, scale, flip and rotation are applied during data synthesis process. The transformed tumor or organs are pasted onto randomly selected background images.}
\label{fig:BraTS-18}
\end{figure*}

\subsection{Results}
Our main results are shown in Table \ref{Brats-18} and Table \ref{synapse}. We compare our results on different exemplar numbers with the results tested on SAM (zero-shot), SAMed, and Full Set for both datasets, where Full Set represents the results of our model on the full training dataset without synthesized data. Notably, the SAM that we use in the test stage is configured with ViT-H (Huge), which is based on the huge ViT encoder that yields the best performance, while the others are configured with ViT-B (Base). For SAM, Ours, and Full Set, we provide one point prompt for each class (except the background class) of each slice in testing. These points are selected from those furthest from the margin of test classes. For SAMed, we use the synthesized dataset as Ours and use the default prompt embedding in~\cite{zhang2023customized} which does not require point prompt and investigates autonomous segmentation. Total Nums in the Exemplar Nums column of all tables means using all images in train sets without synthesized data. Especially, for the Full Set, since we use the full set training set which is not synthesized and we keep the same training pipeline as ours, this method can be treated as the pseudo upper bound. The HD in all table headers represents the HD95. 

\begin{figure*}[ht]
\begin{center}
\includegraphics[width=\linewidth]{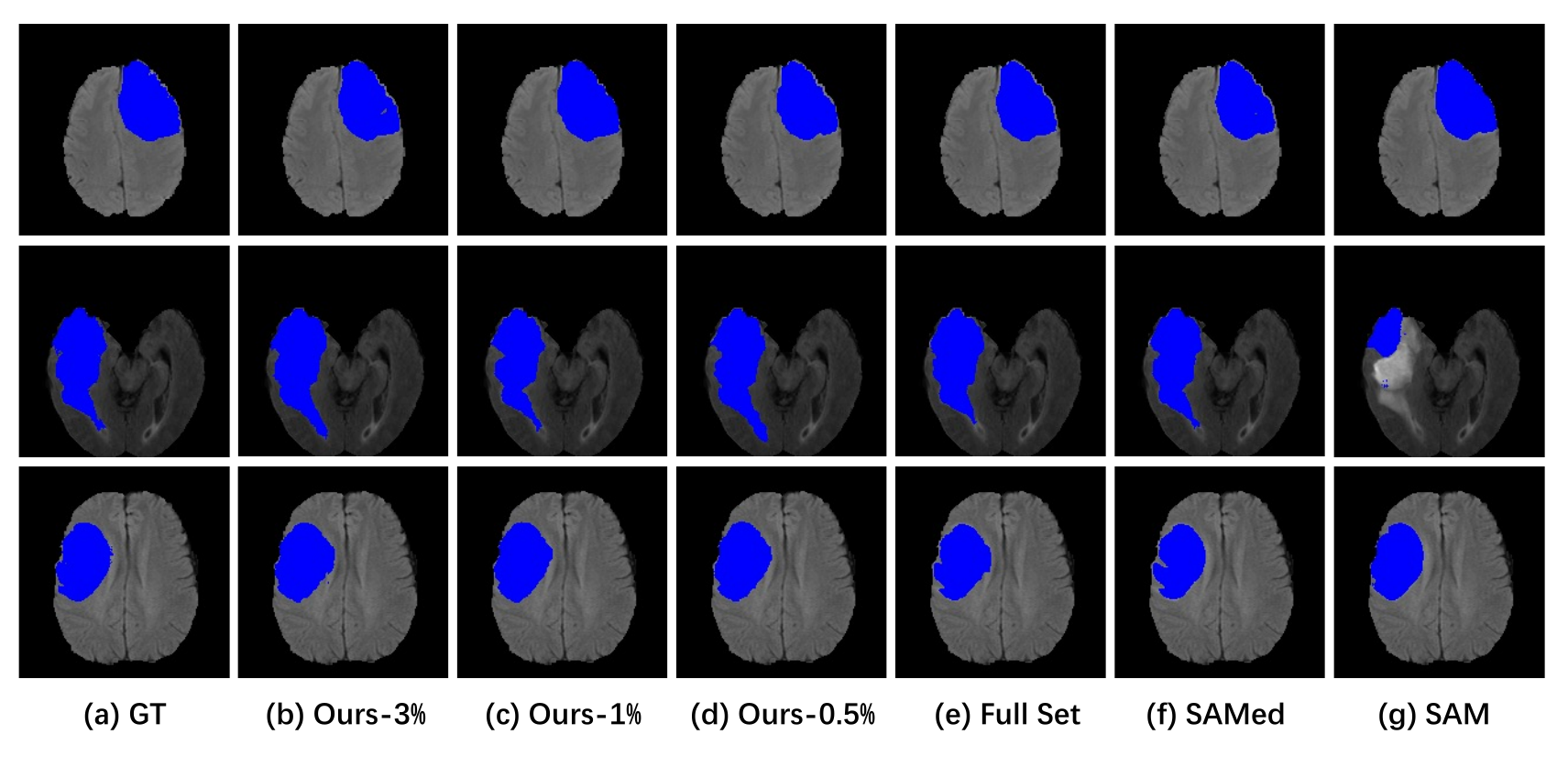}
\end{center}
\vspace{-0.5cm}
   \caption{The qualitative comparisons on the BraTS 2018.}
\label{fig:BraTS-18}
\end{figure*}

\begin{figure*}[!h]
\begin{center}
\includegraphics[width=\linewidth]{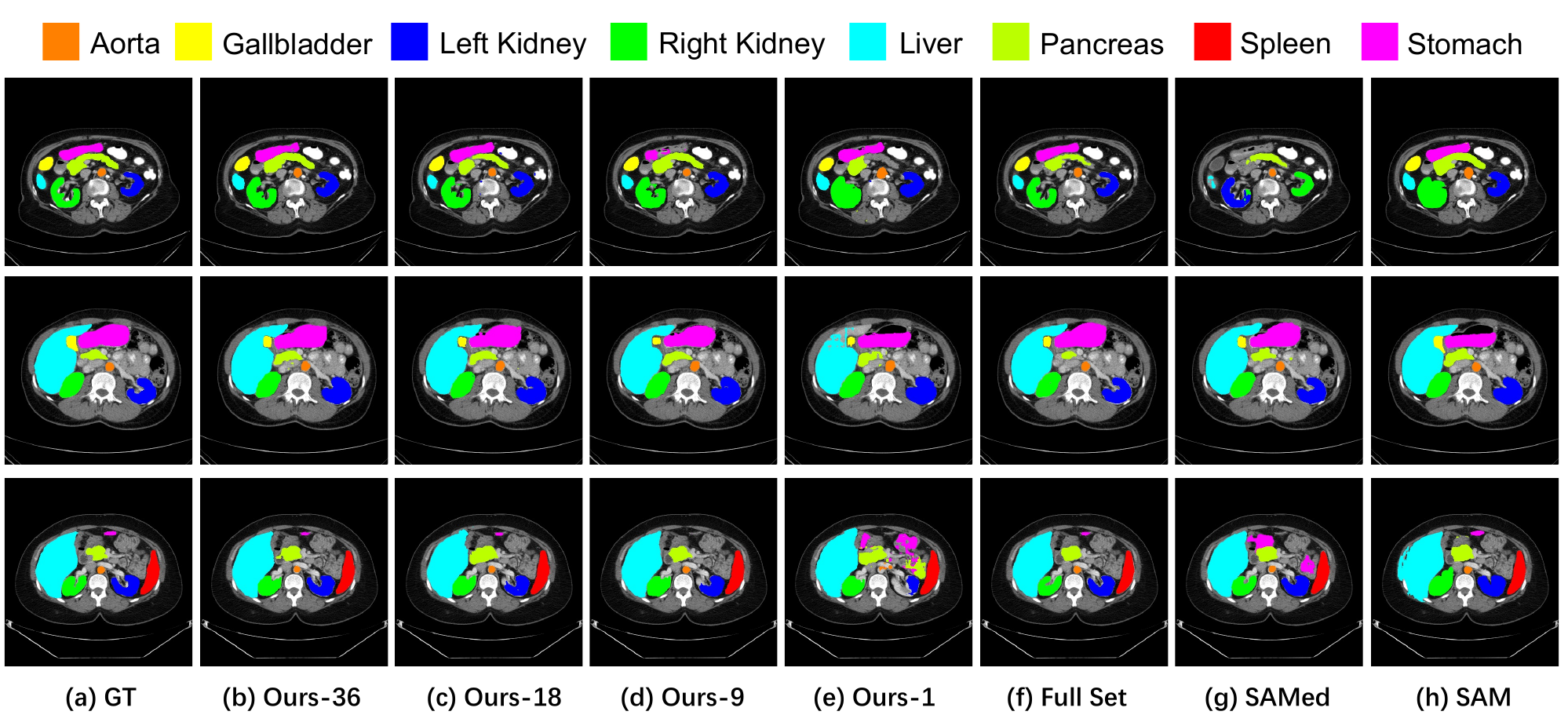}
\end{center}
\vspace{-0.4cm}
   \caption{The qualitative comparisons on the Synapse multi-organ CT dataset.}
\label{fig:Synapse}
\end{figure*}

Table \ref{Brats-18} shows the performance of SAM, SAMed, Ours, and Full Set on the BraTS.
The results tested directly with SAM are poor, indicating the SAM may fail to segment anything in the medical domain. All fine-tuned models achieve much better performances.
Trained on the synthesized dataset based on 0.5\% of the total data, SAMed and Ours achieve rather good test results. Our model outperforms SAMed on HD95 slightly in all test cases. Figure \ref{fig:BraTS-18} gives the qualitative comparison results on the BraTS 2018. 

Table \ref{synapse} shows the results of SAM, Att-UNet, SAMed, Ours, and Full Set on the Synapse. The SAM yields decent results when tested directly on this dataset while SAMed performs poorly when trained on synthesized dataset generated by few exemplars. In contrast, Ours achieves rather good segmentation results with the synthesized dataset using few exemplars. The results of Ours (ViT-B) trained on the dataset generated from only one exemplar is slightly better than SAM (ViT-H) zero-shot performance.
When the number of exemplars comes to 18, the DSC and HD95 in the results are better than those tested on SAM by 8.5\% and 24.06 respectively. The qualitative comparison results are given in Figure \ref{fig:Synapse}. 

Comparing the results on two datasets, we find that point prompts play a more important role in the multi-organ segmentation. We believe that it is because the task of multi-organ segmentation is much harder than the whole tumor segmentation. Though our method is not an automatic solution to medical image segmentation, we attempt to investigate the point prompt setting and demonstrate that our method achieves impressive result with few exemplars. Especially for datasets lacking annotations, our method provides a cost-effective solution to achieve satisfactory segmentation results.

\begin{table*}[t!]
\centering
\caption{Quantitative comparison between w/o and w/ data synthesis on the BraTS.}
\scriptsize
\begin{tabular}{c|c|cc}
\toprule
\textbf{Methods} & \textbf{Exemplar Nums} & \textbf{DSC} $\uparrow$ & \textbf{HD} $\downarrow$\\
\midrule
\midrule
\multirow{3}[1]{*}{w/o Data Synthesis} & 75 (0.5\%) & 68.14 & 17.85\\
& 150 (1\%) & 81.76 & 18.43\\
& 450 (3\%) & 81.02 & 11.14\\
\midrule
\multirow{3}[1]{*}{w/ Data Synthesis} & 75 (0.5\%) & 82.78 & 14.92 \\
& 150 (1\%) & 83.4 & 10.03 \\
& 450 (3\%) & 83.07 & 16.94 \\
\bottomrule
\end{tabular}
\label{Ablation-Brats-18}

\vspace{0.6cm}
\centering
\caption{Quantitative comparison between w/o and w/ data synthesis on the Synapse.}
\footnotesize
\resizebox{\textwidth}{!}{
\begin{tabular}{c|c|cc|cccccccc}
\toprule
\textbf{Methods} & \textbf{Exemplar Nums} & \textbf{DSC} $\uparrow$ & \textbf{HD} $\downarrow$ & \textbf{Aorta}& \textbf{Gallbladder}& \textbf{Kidney(L)}& \textbf{Kidney(R)}& \textbf{Liver}& \textbf{Pancreas}& \textbf{Spleen}& \textbf{Stomach} \\
\midrule
\midrule
\multirow{3}[1]{*}{w/o Data Synthesis} & 9 (one per two volumes) & 78.36 & 24.11 & 85.67 & 52.14 & 87.17 & 85.64 & 86.46 & 64.11 & 88.72 & 76.99 \\
&18 (one per volume) & 79.01 & 20.37 & 86.69 & 52.02 & 87.68 & 86.00 & 88.20 & 63.56 & 90.44 & 77.46 \\
&36 (three per volume) & 80.82 & 18.22 & 89.60 & 50.51 & 87.36 & 87.23 & 92.29 & 65.94 & 92.44 & 81.20 \\
\midrule
\multirow{3}[1]{*}{w/ Data Synthesis} & 9 (one per two volumes) & 79.08 & 21.62 & 88.75 & 55.76 & 88.35 & 84.11 & 89.76 & 61.26 & 91.27 & 73.41 \\
&18 (one per volume) & 83.04 & 16.84 & 89.18 & 71.33 & 89.20 & 86.46 & 92.55 & 64.20 & 90.52 & 80.84 \\
&36 (three per volume) & 84.23 & 11.86 & 88.31 & 69.91 & 90.43 & 88.57 & 94.82 & 65.17 & 91.36 & 85.24 \\
\bottomrule
\end{tabular}
}
\label{Ablation-synapse}
\end{table*}

\subsection{Ablation Studies}
Furthermore, we compare the performances of training on synthesized datasets with those of training on simply exemplars. The results are displayed in Table \ref{Ablation-Brats-18} and Table \ref{Ablation-synapse}. We can see that for both datasets, training on the generated dataset yields better results than training simply with exemplars. This ablation study indicates that on the one hand, a small portion of training data can train the model fairly well, demonstrating SAM has strong learning ability on segmentation. On the other hand, using synthesized data besides exemplars brings obvious improvement to the model performance, shows that exemplar-based data synthesis plays a vital role in improving model performances. The test results using  random point prompts are given in Table \ref{Brats-18-random-point} amd Table \ref{synapse-random-point}.

\begin{table*}[ht]
\centering
\caption{Quantitative results with random point prompts on the BraTS.}
\scriptsize
\begin{tabular}{c|c|cc}
\toprule
\textbf{Methods} & \textbf{Exemplar Nums}& \textbf{DSC} $\uparrow$ &\textbf{HD} $\downarrow$\\
\midrule
\midrule
SAM & - & 44.93 & 55.36 \\
\midrule
\multirow{3}[1]{*}{Ours} & 75 (0.5\%) & 82.88 & 14.49 \\
 & 150 (1\%) & 83.50 & 14.78 \\
 & 450 (3\%) & 83.04 & 14.78 \\
\midrule
Full Set & Total Nums & 85.28 & 7.92 \\
\bottomrule
\end{tabular}
\label{Brats-18-random-point}
\end{table*}

\begin{table*}[ht]
\centering
\caption{Quantitative results with random point prompts on the Synapse.}
\footnotesize
\resizebox{\textwidth}{!}{
\begin{tabular}{c|c|cc|cccccccc}
\toprule
\textbf{Methods} & \textbf{Exemplar Nums} & \textbf{DSC} $\uparrow$ &\textbf{HD} $\downarrow$ & \textbf{Aorta}& \textbf{Gallbladder}& \textbf{Kidney(L)}& \textbf{Kidney(R)}& \textbf{Liver}& \textbf{Pancreas}& \textbf{Spleen}& \textbf{Stomach} \\
\midrule
\midrule
SAM& - & 74.49 & 41.15 & 88.73 & 40.58 & 87.11 & 82.55 & 88.59 & 53.67 & 83.64 & 71.06 \\
\midrule
\multirow{4}[1]{*}{Ours} & 1 (one exemplar) & 72.88 & 25.36 & 82.32 & 43.64 & 83.66 & 83.55 & 84.46 & 50.72 & 89.46 & 50.72 \\
& 9 (one per two volumes) & 76.83 & 16.05 & 88.40 & 52.45 & 87.59 & 82.84 & 87.15 & 57.58 & 88.97 & 69.64 \\
& 18 (one per volume) & 78.53 & 26.74 & 88.03 & 53.61 & 88.50 & 86.27 & 89.08 & 59.08 & 87.64 & 76.06 \\
& 36 (two per volume) & 81.25 & 17.20 & 88.23 & 58.63 & 90.28 & 87.70 & 92.50 & 60.15 & 90.49 & 82.70 \\
\midrule
Full Set & Total Nums & 85.06 & 12.43 & 90.59 & 60.68 & 91.54 & 90.07 & 94.75 & 62.13 & 94.08 & 88.64 \\
\bottomrule
\end{tabular}
}
\label{synapse-random-point}
\end{table*}

\section{Conclusion}
In this paper, we explore the potential of fine-tuning the Segment Anything Model with few exemplars for medical image segmentation. Our approach, which integrates an exemplar-guided synthesis module and the low-rank-based fine-tuning strategy as the data-level and the model-level attempts, has demonstrated promising results in brain tumor segmentation and multi-organ CT segmentation tasks. The experimental results indicate the feasibility of achieving a balance between accuracy and annotation labor, thereby offering a cost-effective solution for medical image segmentation. Looking ahead, we believe that the effective utilization of limited labeled data remains an open problem, particularly in the current era where foundational models present both opportunities and challenges. Furthermore, the integration of a small number of labeled exemplars with a large amount of unlabeled data is an area that deserves further investigation. We hope our attempts present an important initial step towards the practical application of pre-trained models in the medical domain. 

\section*{Acknowledgement} The work described in this paper was partially supported by grants from the National Natural Science Fund (62201483) and the Research Grants Council of the Hong Kong Special Administrative Region, China (T45-401/22-N).

\bibliographystyle{splncs04}
\bibliography{reference}

\end{document}